# Clustering Time-Series Energy Data from Smart Meters

Alexander Lavin[1], Diego Klabjan[2]

**Abstract**

Investigations have been performed into using clustering methods in data mining time-series data from smart meters. The problem is to identify patterns and trends in energy usage profiles of commercial and industrial customers over 24- hour periods, and group similar profiles. We tested our method on energy usage data provided by several U.S. power utilities. The results show accurate grouping of accounts similar in their energy usage patterns, and potential for the method to be utilized in energy efficiency programs.

---

[1] lavin@cmu.edu
[2] Department of Industrial Engineering and Management Sciences, Northwestern University, Evanston, IL; d-klabjan@northwestern.edu

# 1. Introduction

Energy efficiency programs range in their design, but many can benefit from accurate comparisons of participants. Qualitative characteristics such as building type, location, size, etc. do not suffice in grouping participants for comparison because they do not capture their energy usage tendencies, and the data is difficult to obtain. Quantitatively, energy data alone does not provide enough depth to identify both strengths and shortcomings in participants' energy efficiency. Yet data mining can provide insight to a participant's energy usage tendencies.

An energy usage profile – a series of energy (kWh) as a function of time – is time-series data that can reveal a lot about energy efficiency. A time-series may consist of many individual data values, but can also be viewed as a single object, or in this case, an energy profile. Clustering can be used to measure the similarity between these objects, and group them based on energy usage. The goal in clustering time-series data is to organize the data into homogeneous groups, maximizing the similarity and dissimilarity within and between groups, respectively.

In our study, data was supplied by two power utilities, where each participant is a commercial or industrial (C&I) individual customer account. The energy usage data of these accounts provide daily profiles at granularity of one hour. The two utilities use smart meters for select accounts. Clustering accounts based on these profiles allows us to compare and contrast those with similar energy usage tendencies. Previous studies on time-series clustering of energy profiles either focused on clustering by date of an individual customer, or on residential accounts with the goal of customer segmentation. Our accounts are C&I and the goal of clustering is to identify 'peer accounts' and potentials for energy efficiency. A secondary goal is to identify open and close hours of business. Through clustering, each cluster has an average profile. It is not difficult to 'manually' identify open and close hours of such average profiles (there are as many as there are clusters). Individual participants in energy efficiency programs can be made aware of positives and negatives of their energy use *in comparison to those most similar to themselves in energy use*.

The remainder of the paper is organized as follows. Section 2 discusses several previous works in time-series clustering with relevant applications. Section 3 covers the design of our clustering method. Section 4 presents the case study, and gives more details about our handling of raw data and presentation of data mining results. Section 5 wraps up this paper by discussing the potential for this clustering method in energy efficiency applications, and further investigations.

# 2. Background

Clustering methods have been employed, albeit sparingly, as a statistical technique in mining time-series data. Several studies have been done specifically with energy usage data over time. A survey done by Liao [1] summarizes previous work in clustering time-series data. He discusses many clustering methods across a wide array of applications. Next we outline work on time-series clustering of power data.

The data mining work of Košmelj and Batagelj [2] is of interest because of several similarities to our work. First, their application was commercial energy consumption over time. Second, their clustering problem also deals with multivariate data and equal length intervals. However, their study differs from ours because the data they analyze was annual consumptions of energy sources, not short intervals of energy usage. They employed a relocation clustering procedure, modified to incorporate the time dimension with time-dependent linear weights. The distance measure was Euclidean. To identify good clusters they used the generalized Ward evaluation criterion function.

J.J. van Wijk and E.R. van Selow [3] used the root mean square distance function in an agglomerative hierarchical approach in clustering time-series data for two cases also at daily intervals: power demand and employee attendance. They also cluster energy use in daily intervals. The results were delivered with calendar-based visualization, which supported plans to cluster by calendar day. They considered combining time scales, i.e. weekdays of February, but rejected the idea anticipating difficulty in extracting information from so many scenarios. The main intent of their study is to find dates in a year with similar patterns for a single individual account. This is a stark contrast with our goal of clustering customers over fixed dates, i.e., all weekdays in a given month.

Smith et al. [4] investigated the potential of clustering as a data mining technique in segmenting residential customers for energy efficiency programs. The data was filtered to represent residential customers in zones of high daytime energy use, for weekdays, and for summer months June and July. The data was further filtered to exclude households of "high entropy," or chaotic distribution of energy profiles from day-to-day. Average load shapes were clustered via k-means with a Euclidean distance measure, partitioning into six distinct energy profiles by their main peaks: afternoon, evening, dual, daytime, night, and morning. While we focus on C&I customers, they focus on residential participants. They also narrow the focus to a very specific geographic zone and timeframe. We applied our clustering method to a wide scope of participants and scenarios.

## 3. Cluster Analysis

### 3.1 Approach

In order to achieve the goal of our clustering – to merge energy profiles into clusters where profiles in a common cluster are more similar than those in other clusters – we use a standard partitioning method. Of the major clustering approaches, partitioning was chosen over hierarchical, density-based, grid-based, model-based, and constraint-based because partitioning methods are distance-based. This is an ideal feature for our application because we aim to compare and contrast individual energy profiles, which can be accomplished with distinct distance measures providing a tangible dissimilarity measure. The clustering method maintains its "structural" approach by accumulating a measure of dissimilarity over the time-series interval, giving a total distance for a given data profile.

The clustering method k-means was selected as a standard heuristic partitioning method. This was identified as the grouping of energy use profiles is based on the similarity to a mean profile. This is desired because mean profiles can be used in energy efficiency comparisons.

With the comparison of individual energy profiles as our goal, normalization of the data is necessary. Accounts may mirror each other in energy usage tendencies, but not in their kWh throughout the day because of differences in the underlying environment. Normalization scales the full dataset of all accounts such that the energy profiles are clustered based on their "structural" similarities. The resulting plots and comparisons reflect true strengths and weaknesses in energy efficiency.

In a dataset of $N$ data tuples, $k$ partitions of the data are created. Each partition represents a cluster where each cluster has at least one data object and $k \leq N$. In our problem of clustering time-series data, each data tuple is an energy profile of an average 24-hour period. Thus, within each tuple there are $n$ intervals with individual energy measurements covering the 24-hour period – i.e. if the length of an interval is 15 minutes, there are 96 data entries in a single energy profile, or $n = 96$ in each data tuple.

### 3.2 Distance

Partitioning clustering algorithms are distance-based, relying on a dissimilarity measure to assign data objects to their optimal cluster. This measure, the distance, determines the similarity between two data tuples. To allocate energy profiles to their appropriate clusters, we use a dissimilarity measure to calculate the distance between each possible combination of two energy profiles in the dataset. The two time-series in a given calculation are sampled on the same interval and of the same length. There are many distance measures to consider implementing.

Let $a_i$ and $b_j$ each be an $n$-dimensional vector. The difference of squares measure is computed as

$$d_{dos} = \sum_{k=1}^{n}(a_{ik} - b_{jk})^2.$$

The Euclidean distance follows simply as

$$d_E = \sqrt{d_{dos}}.$$

The average geometric distance, known as the root mean squared distance, is

$$d_{rms} = d_E/n.$$

Alternatively, a normalized distance function may be used. Here the series $a$ and $b$ are normalized by dividing by the maximum value in each respective sequence,

$$d_{nm} = \sqrt{\sum_{k=1}^{L}\left(\frac{a_{ik}}{a_i^{max}} - \frac{b_{jk}}{b_j^{max}}\right)^2}/n$$

where $a_i^{max} = \max_k a_{ik}$ and $b_j^{max} = \max_k b_{jk}$.

### 3.3 Methodology

We use a calendar-based approach to organizing the data. For a given account, this yields an energy profile representative for each day of the year. More details on this averaging can be found in our case studies discussion in Section 4.

The clustering process starts with a predetermined number of $k$ clusters and days of the calendar year. The distance $d$ is computed, using one of the aforementioned metrics, between every possible pair of energy profiles in the dataset. The calculated dissimilarities are stored in a data table. The *k*-means clustering algorithm is then run. Plots of energy versus time are then made to analyze the results. Visualization of energy profiles within clusters, as well as comparing the average profiles of clusters, is a very good method of deciding if $k$ or $d$ needs an adjustment. Trial runs over different interval dates either reinforce or repudiate an option of $k$ or $d$.

## 4. Case Studies

Raw energy usage data was provided by two power utilities located in the Midwest region of the United States. The accounts are commercial businesses, of which we know their categories – e.g., public schools and small cafés – and are on the order of 1,000 for each utility. The datasets cover several years post-2007, and the smart meter measurements are hourly.

The data was stored in Apache Cassandra database management system. All queries to the data were made from Cassandra. The clustering calculations – distance measures and *k*-means – were done in R statistical software environment. Results were stored back in Cassandra. By using Cassandra, a single clustering can be obtained in minutes, while using a standard relational database takes more than 10 hours (we tried the community version of PostgreSQL).

We tried the various distance measures introduced in Section 3. Extensive computational experiments revealed that they do not produce drastically different clusters. We settled for $d_E$ since it is the measure frequently used in other applications.

### 4.1 Data Cleansing

Raw energy usage data in most cases has gaps, which can range anywhere from hours to months, and estimations of true meter readings. Our approach in data cleansing was to fill the gaps of missing data via linear interpolation or averages, and to identify and mark estimated data.

When the missing data covered several hours or less, simple linear interpolation using data before and after the interval sufficed. However, it was often the case that missing data would cover a greater time frame – e.g. 13 hours or several days. With several years of data for each account at our disposal, we averaged the corresponding energy values from other years in the dataset. For example, the missing values from 8:00 January 2[nd] to 8:00 January 9[th] for a given year would be filled with values averaged from other years' 8:00 January 2[nd] to 8:00 January 9[th].

The estimated data was identified by searching for energy usage values repeated consecutively. The number of consecutive repetitions must reach a threshold value in order to be deemed an estimation. If the flagged data covers relatively short intervals, it is replaced via the aforementioned interpolation method. However, if the interval is of considerable length, the data is removed from the dataset. Furthermore, if an account's data is found to have many instances of estimation such that it compromises the validity of the data, the account is removed from the case study. The final cleansing step is normalization, as aforementioned, to allow for comparison of energy usage tendencies.

**4.2 Visualization**

An individual energy profile is a plot of time vs. energy with the domain 0-24 hours and the range greater than or equal to zero kWh. Our raw data from the utilities was given in 15-minute intervals, which we summed to hourly intervals for faster computing, still yielding reasonably smooth plots over a 24-hour period.

For a single cluster, we produced a plot of all energy profiles in the cluster, as well as a plot of the average energy profile. The former allows us to identify energy efficiency issues of individual accounts. Both plotting schemes help in determining the optimal $k$ and $d$ inputs. Van Wijk and van Selow [3] display power demand as a function of both power and days in a three-dimensional mountain landscape. This illustrates profile differences over many days and may be used to focus on a single account to identify trends and entropy in energy use. Our two-dimensional illustration of the results suffices to demonstrate the clustering method's effectiveness in revealing energy efficiency strengths and weaknesses of accounts.

**4.3 Results**

We focus on a specific commercial business category from one of the power utilities. By traditional qualitative grouping methods, all 821 accounts in the Food Sales category would be initially grouped together for comparison, and then subdivided based on power consumption. Our clustering method, however, divides the Food Sales accounts based on their energy usage patterns.

The results are presented in energy profile plots of each cluster. Within each cluster plot, accounts most similar in energy use with each other can be compared. Hour-by-hour, any data points that stray from the "cluster mean profile" can be identified as either positive or negative contributions to one's energy efficiency.

The plots shown in Figure 1 result from imposing nine clusters ($k$ = 9) and the Euclidean distance measure for June. The clustering algorithm succeeds in grouping accounts most similar in energy usage. Individual deviations from a cluster's mean profile are also observed, highlighting an account's better or worse use of energy than its peers. For instance, Figure 2 focuses on the plot of Food Service accounts open from 4 PM to midnight, or "open at 16 for 8 hours." This cluster contains 48 accounts, or 6% of the total accounts in the category. Most accounts follow the mean tightly, but around 12 PM we can identify energy efficient accounts as those deviating below the mean profile. Others that spike above the profile have room to improve. Informing both parties can be beneficial for an energy efficiency program. The positive reinforcement of good-energy practices can keep businesses energy efficient, and highlighting subpar performers can motivate them to improve their practices for better energy use.

Consider now Figure 3, showing the equivalent cluster, but for weekdays in months nine and ten – September and October. The cluster now contains 38 accounts, or 10 fewer than from June. Ten accounts have been relocated to other clusters. This variability in a cluster's peers highlights an important feature of our clustering method: energy trends vary month-by-month, so we must be able to identify energy efficiency issues on a monthly basis. Clustering for different time frames allows comparison of an account's energy use to its *most similar peers*.

The comparison of Figures 2 and 3 shows a specific instance of a change in energy efficiency. In Figure 2, between the 15 and 20 hour marks, we observe one account straying well below the mean energy profile of the cluster, but in Figure 3 we do not. This can be attributed to a change in energy use by the account between the two sample time periods. Identifying changes in energy use such as this can significantly benefit an energy efficiency program's ability to help its participants.

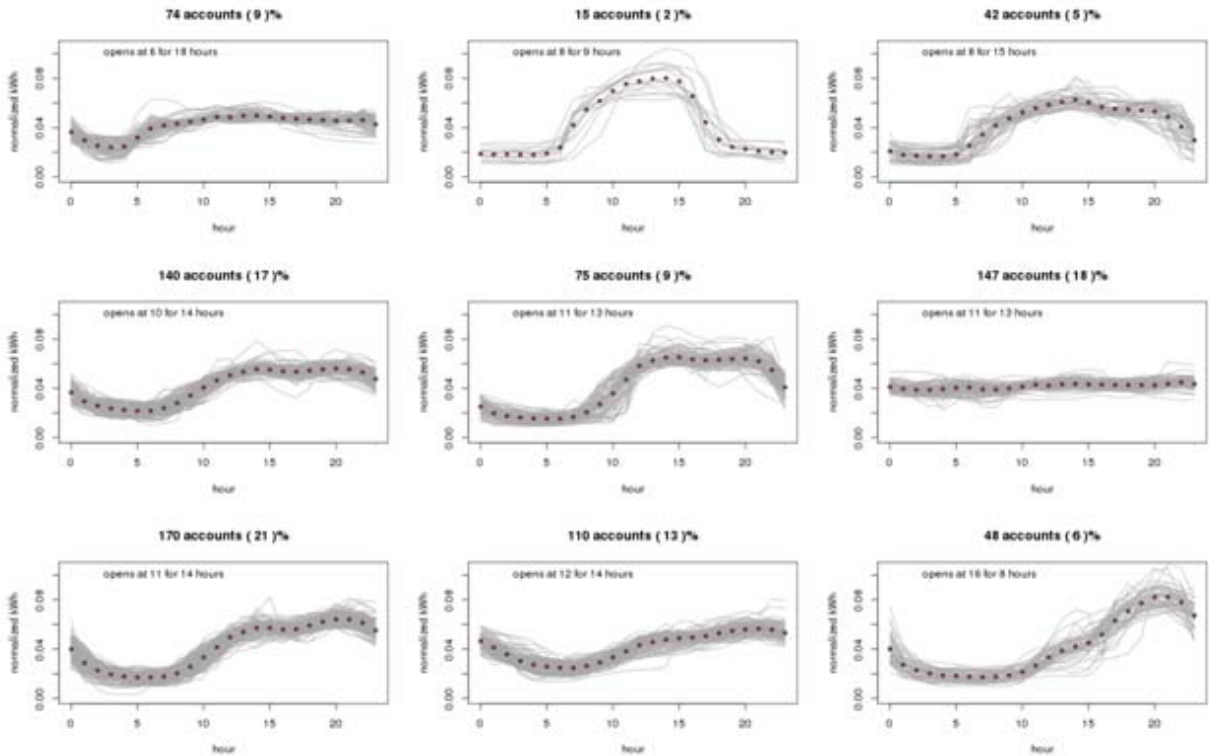

Figure 1. Food Service accounts clustered for weekdays in June.

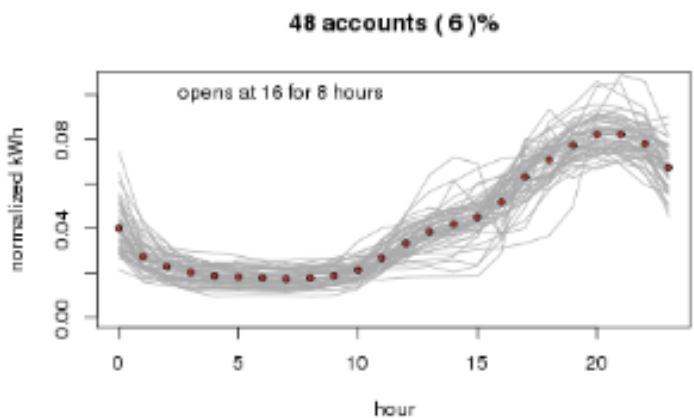

Figure 2. A cluster for Food Service for weekdays in September and October.

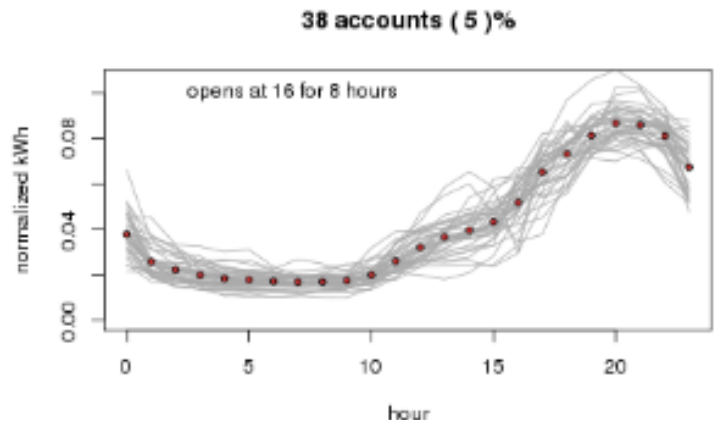

Figure 3. A cluster for Food Service for weekdays June.

# 5. Discussion

Our goal was to use data mining to identify and compare accounts with *truly similar energy use* – following the same power pattern for a given set of days. We have presented a clustering method for time-series data with the intended application of energy efficiency. We have identified that there is potential for the method to identify energy efficiency issues of individual accounts. Further evaluating of clustering parameters may be done by employing an evaluation criterion such as the generalized Ward criterion considered by Košmelj and Batagelj [2]. There are also additional time-series clustering techniques to consider, such as fuzzy c-means shown effective for short time-series data [5].

An issue can arise with our use of normalization in the clustering method. To illustrate the motivation behind normalization, consider an example of two fast food locations, Alpha and Beta, which are franchises of the same restaurant chain. They have the same hours, menu, etc., such that their energy usage patterns are almost identical. The one difference, however, is that Beta leaves the air conditioning unit on maximum during closed hours from 2-5PM, while Alpha turns it down. But Alpha is five times the size of Beta, and thus uses five times the power. Without normalization, Alpha and Beta would not be clustered together. Consequentially we would not notice the slight peak in energy use of Beta from 2-5 PM as a specific instance of energy inefficiency. That is, normalizing the plots of Alpha and Beta allows us to adequately compare their energy use to find any differences in their efficiencies. Without normalization, clusters would group accounts most similar in their *capacities* of energy use, not necessarily their *patterns* of energy use.

In conclusion, we believe the clustering method presented here is useful to identify energy efficiency issues and strengths in individual accounts. The method has potential to be used as a data mining tool in energy efficiency programs and studies; perhaps energy efficiency recommendations and tips can supplement our statistics.

# References

[1] T.W. Liao, "Clustering of time series data – a survey." Pattern Recognition Society 38 (2005). Elsevier Ltd., 2005.

[2] K. Košmelj, V. Batagelj, "Cross-sectional approach for clustering time varying data." J. Classification 7 (1990): pp. 99–109.

[3] J.J. van Wijk, E.R. van Selow, "Cluster and calendar based visualization of time-series data." Proceedings of IEEE Symposium on Information Visualization, San Francisco, CA, October 25–26, 1999.

[4] Smith, Brian A., Arthur Wong, and Ram Rajagopal. "A Simple Way to Use Interval Data to Segment Residential Customers for Energy Efficiency and Demand Response Program Targeting." ACEEE (2012). Web. 18 Nov. 2012.

[5] C.S. Möller-Levet, F. Klawonn, K.-H. Cho, O. Wolkenhauer, "Fuzzy clustering of short time-series and unevenly distributed
sampling points." Proceedings of the 5th International Symposium on Intelligent

Data Analysis, Berlin, Germany, August 28–30, 2003.